\documentclass{article}

\usepackage[preprint]{neurips_2026}

\usepackage[utf8]{inputenc} 
\usepackage[T1]{fontenc}    
\usepackage{hyperref}       
\usepackage{url}            
\usepackage{booktabs}       
\usepackage{amsfonts}       
\usepackage{nicefrac}       
\usepackage{microtype}      
\usepackage{xcolor}         
\usepackage{graphicx}
\usepackage{placeins}
\usepackage{amsmath, amssymb}
\usepackage{tikz}
\usetikzlibrary{arrows.meta, calc, shapes.geometric, decorations.pathreplacing}
\usepackage{etoc}
\usepackage{xcolor}

\newcommand{\ourparagraph}[1]{\noindent\textbf{#1}\ }
\title{Contrast encodes inductive bias: separating slow noise from dynamics in predictive representation learning}
\author{Paarth Gulati\\
Department of Physics, and\\
Initiative in Theory and Modeling of Living Systems\\
Emory University, Atlanta, GA\\
\And
Ilya Nemenman\\
Departments of Physics and Biology, and\\
Initiative in Theory and Modeling of Living Systems\\
Emory University, Atlanta, GA\\
}

\begin{document}

\maketitle

\begin{abstract}
Self-supervised methods that learn representations and predict dynamics fully in the latent space, such as JEPA, have been shown to confuse slowly varying noise with the dynamical signals they aim to capture. Specifically, when noise features remain approximately constant within each trajectory, contrastive predictive objectives preferentially encode these features instead of the true latent variables governing the system. The learned representation then becomes dominated by trajectory-specific noise, so downstream performance degrades with noise strength and does not improve even as the number and duration of training trajectories increase. We argue that this failure is a property of the objective itself, shared by a long line of contrastive predictive objectives that sample negatives across trajectories. To illustrate this generality, we study the failure mode and its remedy in two settings: a standard SimCLR-style JEPA on a synthetic moving-dot dataset, and DySIB, a recently introduced method designed for extracting physically interpretable representations of dynamics, on movies of a rigid-body pendulum. When negatives are instead sampled within a single trajectory, the slow noise can no longer distinguish frames within that trajectory, removing the predictive shortcut. Training one encoder simultaneously on many such trajectories then forces it to encode the variables relevant for the dynamics, with longer trajectories yielding better representations even for strong slow noise. Our results point toward principles for designing contrastive predictive objectives in dynamical representation learning, especially for physical systems with noisy experimental observations.
\end{abstract}

\etocdepthtag.toc{mtoc}

\section{Introduction}

Latent-variable models that learn compact representations of high-dimensional observations and predict their dynamics entirely in the latent space, including Joint Embedding Predictive Architectures (JEPAs) and others, have become a central tool for self-supervised learning of physical and embodied systems~\citep{assran2023self,bardes2024revisiting,assran2025vjepa2,balestriero2025lejepa,maes2026leworldmodel,wang2022denoised}. The promise is that an encoder trained in this way will discover the variables that actually govern the system, without supervision and from raw data alone, and that it will do so with substantially less data than self-supervised approaches that predict in the data space~\citep{martini2024dataefficiency,vanassel2025jointembedding}. In practice, what is most predictable in a data stream is often not what we want a representation to capture. Realistic recordings, including the experimental videos used to learn physically interpretable descriptions of dynamics~\citep{chen2022automated,schmitt2024machine,gurevich2024learning,farrell2023inferring,krieger2025interpretable,daniels2019automated,yu2025dusty}, contain many forms of noise across a range of timescales. The most difficult to handle are noises that change little for the duration of a trajectory $T$ and vary only across trajectories, such as background textures, lighting, and dust on the lens. Predictive objectives focus on such noises since their across-trajectory variation supplies the entropy that prediction requires, and their within-trajectory persistence makes the future easy to predict. The variables that actually carry the dynamics or the consequences of actions change with time and are much harder to learn. Thus, a latent representation organized around what is easiest to predict ends up encoding trajectory identity, and the dynamics of the system are lost in the noise.

Similar problems have been observed for a wide class of video self-supervised learning (SSL) methods, where they have been called background bias, shortcut learning, or over-reliance on static features~\citep{wang2021removing,huang2021self,ding2022motion,thoker2023tubelet,kowal2022deeper}. Several remedies have been proposed, typically augmenting positive and negative pairs with image transformations that suppress the static content. The same failure has been flagged for predictive objectives in the JEPA family by~\citet{sobal2022joint}, and analogous fixes have begun to appear~\citep{yu2025auxiliary,nam2026causaljepa,toso2026learning}. Theoretical analyses of related scenarios show that representation learning from video alone under temporally correlated exogenous noise can be exponentially hard~\citep{misra2024principled,efroni2022provable}, and that the idealized predictive loss is minimized when the encoder represents the slowest features of the dynamics~\citep{ruizmorales2025koopman}. It remains unclear whether the failure is a fundamental property of predictive objectives that pair pasts and futures of dynamics more broadly, or whether the right choice of objective resolves it.

We approach this question by systematically studying two contrastive predictive architectures on simulated and experimental data. The first is the SimCLR-predictor of~\citet{sobal2022joint}, evaluated on a simulated dataset, where each trajectory is a stochastic walk of a single dot, with noise added to every pixel of every frame. The second is DySIB~\citep{martini2026dysib} (Dynamical Symmetric Information Bottleneck), applied to experimental video of a rigid pendulum. This recently introduced information-bottleneck encoder~\citep{tishby2000information,friedman2001multivariate,alemi2017deep,abdelaleem2025deep} compresses both past and future of the dynamics while preserving the predictive information~\citep{bialek2001predictability} between them in the latent space. In each case we add noise of varying amplitude $\sigma$ and correlation time $\tau$ to the frames, so that the nuisance can range from independent frame-to-frame fluctuations ($\tau \to 0$) to a fixed pattern that varies only across trajectories ($\tau \to \infty$). The two methods differ substantially, so that studying them together separates failures intrinsic to contrastive predictive learning from those specific to implementations.

We show that in both architectures the standard contrastive predictive objective fails similarly. As $\tau$ approaches the trajectory length, the encoder organizes the latent space around the trajectory-specific nuisance, and probes can no longer recover the dynamical variables. The shared failure across architectures shows that the source of the problem in the contrastive predictive objective itself, independent of other details. The mechanism can be traced to the explicit focus on slow features in a long line of predictive objectives on which modern methods are built~\citep{wiskott2002slow,oord2018representation}. We show that a modification of the contrast structure resolves this failure. The standard objective uses across-trajectory negatives (ATN). We replace these with within-trajectory negatives (WTN), so that the slow nuisance pattern is shared across positives and negatives and cannot be exploited as a predictive shortcut. Provided the trajectory traverses a wide enough region of the system's phase space, positive pairs of nearby pasts and futures lie closer in latent space than negatives drawn from more distant points on the same trajectory, and the objective focuses on features that change with the dynamics, recovering the true dynamical degrees of freedom even under strong slow noise, with longer trajectories systematically reducing the probe error.

Thus, the contrast structure of the objective controls both the failure and its fix. A nuisance shared by positives and negatives gives the encoder nothing to encode, while one that differs between them becomes a predictive shortcut. In dynamics-paired predictive learning, contrast design is then an analogue of augmentation design in image self-supervised learning, with both shaping what the encoder discards~\citep{saunshi2022understanding,vonkugelgen2021self,wang2020understanding,tian2020makes,xiao2021what}. We hope to elevate the structure of the contrast in predictive objectives to a design choice, on a par with the choice of architecture, augmentations, or the form of the loss.

\ourparagraph{Contributions.} This work (i) reproduces the failure mode of \citet{sobal2022joint}, originally demonstrated at the fixed-noise limit, and extends it to intermediate correlation times and to a second distinct architecture, locating the source of the problem in the objective itself; (ii) maps the failure across the parameter space $(\sigma, \tau, T)$, where the baseline fails with increasing noise correlation time while our fix struggles only at $\tau \approx T$; (iii) shows that drawing negatives from within the same trajectory allows to recover the dynamical variables, with the peak probe error decreasing as trajectory length grows; and (iv) frames the contrast structure of a predictive objective as an inductive bias.

\begin{figure}[t]
  \centering
  \centering
  \resizebox{\linewidth}{!}{\input{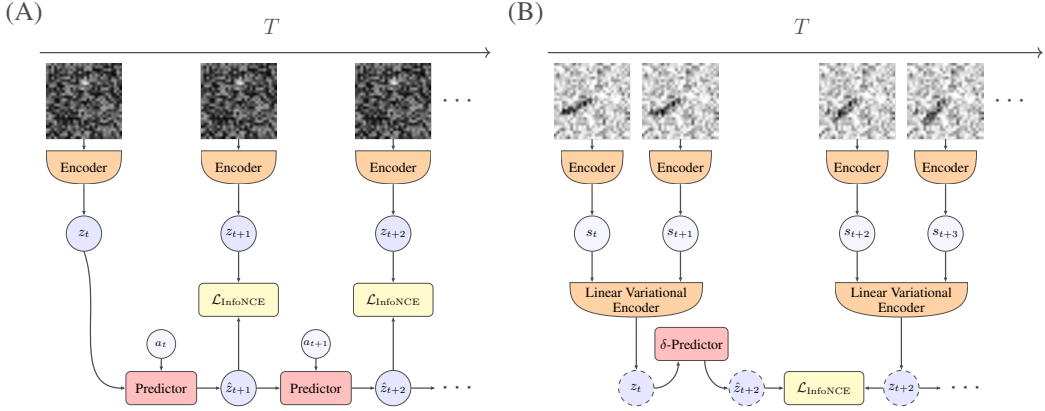}}
  \caption{\textbf{Datasets and architectures.} The two (dataset, architecture) pairs studied here; both panels show example frames from length-$T$ trajectories with added temporally correlated noise. \textbf{(A)} SimCLR-predictor~\citep{sobal2022joint} on a synthetic moving-dot dataset, with an action-conditioned predictor that maps the encoded current frame and action to the encoded next frame. \textbf{(B)} DySIB~\citep{martini2026dysib} on experimental video of a rigid pendulum~\citep{chen2022automated}, with a $\delta$-predictor mapping between non-overlapping $n_F = 2$ frame windows in a Gaussian latent. Dashed circles denote the variational encoder (reparameterization trick). Both methods, as published, use an InfoNCE loss with across-trajectory negatives (ATN). Here we also explore within-trajectory negatives (WTN), see~\S\ref{sec:setup}.}
  \label{fig:architectures}
\end{figure}

\section{Setup}\label{sec:setup}

\ourparagraph{Problem setting.} We consider a training batch of $N$ trajectories $\{x^{(n)}_t\}_{t=1}^{T}$, $n=1,\ldots,N$, sampled from a possibly larger dataset. Each frame $x^{(n)}_t \in \mathbb{R}^D$ is a high-dimensional observation (a movie frame) of a dynamical system, possibly contaminated with temporally correlated noise whose structure we control (see below). The encoder $\phi$ takes a window of $n_F$ consecutive frames, $z^{(n)}_t = \phi(x^{(n)}_{t-n_F+1}, \ldots, x^{(n)}_t) \in \mathbb{R}^{k_z}$, and is trained against a contrastive predictive objective that aligns $z^{(n)}_t$ with its prediction $\hat{z}^{(n)}_{t}$ derived from the context $n_F$ frames before. The choice of negative-sample distribution is treated as a design knob below: across-trajectory negatives (ATN) versus within-trajectory negatives (WTN). At evaluation, the encoder is frozen and a low-capacity probe is trained to predict a known dynamical variable $y^{(n)}_t$ from $z^{(n)}_t$, so that probe accuracy reflects the representation rather than the probe's own capacity.

\ourparagraph{Datasets and architectures.} Figure~\ref{fig:architectures} shows the two (dataset, architecture) pairs we study, both taken from prior work, with added noise (see below) as the only modification; see Appendix~\ref{app:dataset_architectures} for details. The first (Fig.~\ref{fig:architectures}A) is the SimCLR-predictor of~\citet{sobal2022joint} applied to their synthetic moving-dot dataset, in which each trajectory is a random walk of a single dot, with each step set by a random action $a_t \in \mathbb{R}^2$, and noise added to every pixel. A per-frame encoder ($n_F = 1$) maps each frame $x_t$ to a latent $z_t = \phi(x_t)$. Starting from $\hat{z}_1 = z_1$, an action-conditioned predictor $f$ autoregressively rolls out latents $\hat{z}_t = f(\hat{z}_{t-1}, a_{t-1})$ for $t = 2, \ldots, T$, and an InfoNCE~\citep{oord2018representation} loss aligns each $\hat{z}_t$ with the encoded $z_t$.
The second (Fig.~\ref{fig:architectures}B) is DySIB~\citep{martini2026dysib} applied to a public dataset of experimental videos of a rigid pendulum swinging under gravity~\citep{chen2022automated}. A single per-frame encoder $s_t = g(x_t)$ (with shared weights across past and future windows) is followed by a linear variational encoder $h$. Together, $g$ and $h$ map a window of $n_F = 2$ consecutive frames to a Gaussian latent $z_t = h(s_{t-n_F+1}, \dots, s_t)$. Then a $\delta$-predictor maps the past-window latent $z_{t-n_F}$ to a prediction of the next non-overlapping future-window latent $\hat{z}_{t}=z_{t-n_F}+\delta(z_{t-n_F})$. The contrastive InfoNCE loss again aligns the prediction with $z_{t}$. DySIB also includes a regularizer on the latent, but it is weighted at $1/100$ of the predictive term in our experiments and contributes negligibly.
The simulated dot offers full control over dynamics and noise; the recorded pendulum confirms the conclusions on a real physical system with a directly visualizable 2-D state (angle and angular velocity).

\ourparagraph{Noise model.} We add to each frame a slowly varying nuisance whose temporal structure we control. The noise field follows an Ornstein--Uhlenbeck recursion in time, with independent processes at every pixel and every trajectory,
\begin{equation}
    \eta_t = \mu + \alpha\,(\eta_{t-1} - \mu) + \sqrt{1-\alpha^2}\,(\xi_t - \mu), \qquad 0 \le \alpha \le 1,
\end{equation}
where the $\xi_t$ are i.i.d. with mean $\mu$ and $\eta_0 \sim \xi$. To check robustness to the noise distribution, we draw $\xi_t \sim \mathrm{Unif}(0,1)$ ($\mu = 1/2$) for the moving dot (matching~\citet{sobal2022joint}) and $\xi_t \sim N(0,1)$ ($\mu = 0$) for the pendulum, both spatially uncorrelated (\citet{sobal2022joint} demonstrate that spatial structure is irrelevant for the failure mode). The noiseless frame $\hat{x}_t$ is normalized to $[0, 1]$, and the noisy frame is $x_t = \hat{x}_t + \sigma\,\eta_t$. For the pendulum we clip the result to $[0,1]$, while for the moving dot we leave it unclipped to match~\citet{sobal2022joint}. The amplitude $\sigma$ and correlation time $\tau \equiv -1/\ln\alpha$ control the scale and timescale, with autocovariance $\langle\eta_t\eta_{t'}\rangle - \mu^2 \propto e^{-|t-t'|/\tau}$. At $\tau \to 0$ ($\alpha = 0$) the noise is independently sampled at every frame (\emph{changing noise}). At $\tau \to \infty$ ($\alpha = 1$) a single pattern is fixed within a trajectory but varies across trajectories (\emph{fixed noise}), the regime where standard predictive objectives fail. Intermediate values of $\tau$ have not been tested previously to our knowledge, and their exploration is one of our contributions. Since the numerical values of $\sigma$ are not directly comparable across settings, we use distinct symbols $\sigma_U$ and $\sigma_G$ and report results for each case separately.

\ourparagraph{Standard predictive objective: Across-trajectory negatives.}
Both architectures are trained with an InfoNCE contrastive loss~\citep{oord2018representation}. A batch consists of $N$ trajectories of $T$ time points each, with trajectory indices $\mathcal{B}=\{1,\ldots,N\}$ and time indices $\mathcal{T}=\{1,\ldots,T\}$. For each predicted future latent $\hat z_t^{(n)}$, the matched positive is the corresponding future encoding $z_t^{(n)}$, and the per-anchor and batch losses are
\begin{equation}
    \ell_t^{(n)} = -\log\frac{\exp s(\hat z_t^{(n)}, z_t^{(n)})}{\exp s(\hat z_t^{(n)}, z_t^{(n)}) + \sum_{z^- \in \mathcal{N}_t^{(n)}} \exp s(\hat z_t^{(n)}, z^-)}, \quad \mathcal{L} = \frac{1}{NT}\sum_{n\in\mathcal{B},\, t\in\mathcal{T}} \ell_t^{(n)},
    \label{eq:infonce}
\end{equation}
where $s$ is a similarity score (cosine similarity for the SimCLR-predictor, predictive Gaussian log-density for DySIB; see Appendix~\ref{app:dataset_architectures} for details) and $\mathcal{N}_t^{(n)}$ is the negative set associated with the anchor.\footnote{\citet{sobal2022joint} use a symmetrized variant of $\ell_t^{(n)}$ while DySIB does not; the difference does not affect our analysis.} The standard choice for both architectures takes across-trajectory negatives (ATN),
\begin{equation}
    \mathcal{N}_{{\rm ATN},t}^{(n)} = \left\{ z_t^{(n')} : n' \in \mathcal{B},\ n' \neq n \right\},
    \label{eq:atn}
\end{equation}
so the matched future encoding $z_t^{(n)}$ is contrasted against future encodings at the same time index from the other trajectories in the batch.

\ourparagraph{Within-trajectory negatives.}
The within-trajectory variant (WTN) keeps the InfoNCE form, Eq.~\eqref{eq:infonce}, unchanged but replaces the negative set with samples drawn from the same trajectory at other times,
\begin{equation}
    \mathcal{N}^{(n)}_{{\rm WTN},t}
    =
    \left\{
    z^{(n)}_{t'} : t' \in \mathcal{T},\ t' \neq t
    \right\}.
    \label{eq:wtn}
\end{equation}
The slow nuisance pattern is now shared by the positive and negative samples within the same trajectory, so it can no longer distinguish the matched pair from the negatives. WTN requires each trajectory to span a representative portion of the system's phase space within its $T$ frames, so that negatives drawn from the same trajectory define a meaningful contrast.

\ourparagraph{Evaluation metrics.}
To evaluate representation quality, we train low-capacity probers to predict known dynamical targets $y^{(n)}_t$: the dot position $(x, y)$ for the moving-dot dataset and $(\cos\theta, \sin\theta)$ for the pendulum (the pendulum's angular velocity is also available but we focus on the angle). Since for the moving-dot we use a high-dimensional latent representation, it suffices for the probe to be a linear readout directly from the learned representation. For the pendulum, latent representations are low-dimensional and nonlinearly related to the variables of interest (Fig.~\ref{fig:embedding_baseline_dysib_kz_2}). Thus we use a fixed random-features map followed by a trainable linear readout, which gives a simple nonlinear probe while keeping the trainable part linear. We report the probe RMSE, sweeping the noise amplitude $\sigma$, the correlation time $\tau$, and the trajectory length $T$. As a reference, we compare against a supervised baseline using the same encoder architecture trained directly on $y^{(n)}_t$ with an MSE loss. Train/test splits and prober architectures follow the original publications~\citep{sobal2022joint,martini2026dysib} and are detailed in Appendix~\ref{app:dataset_architectures}.

\begin{figure}[t]
    \centering
    \includegraphics[width=0.8\linewidth]{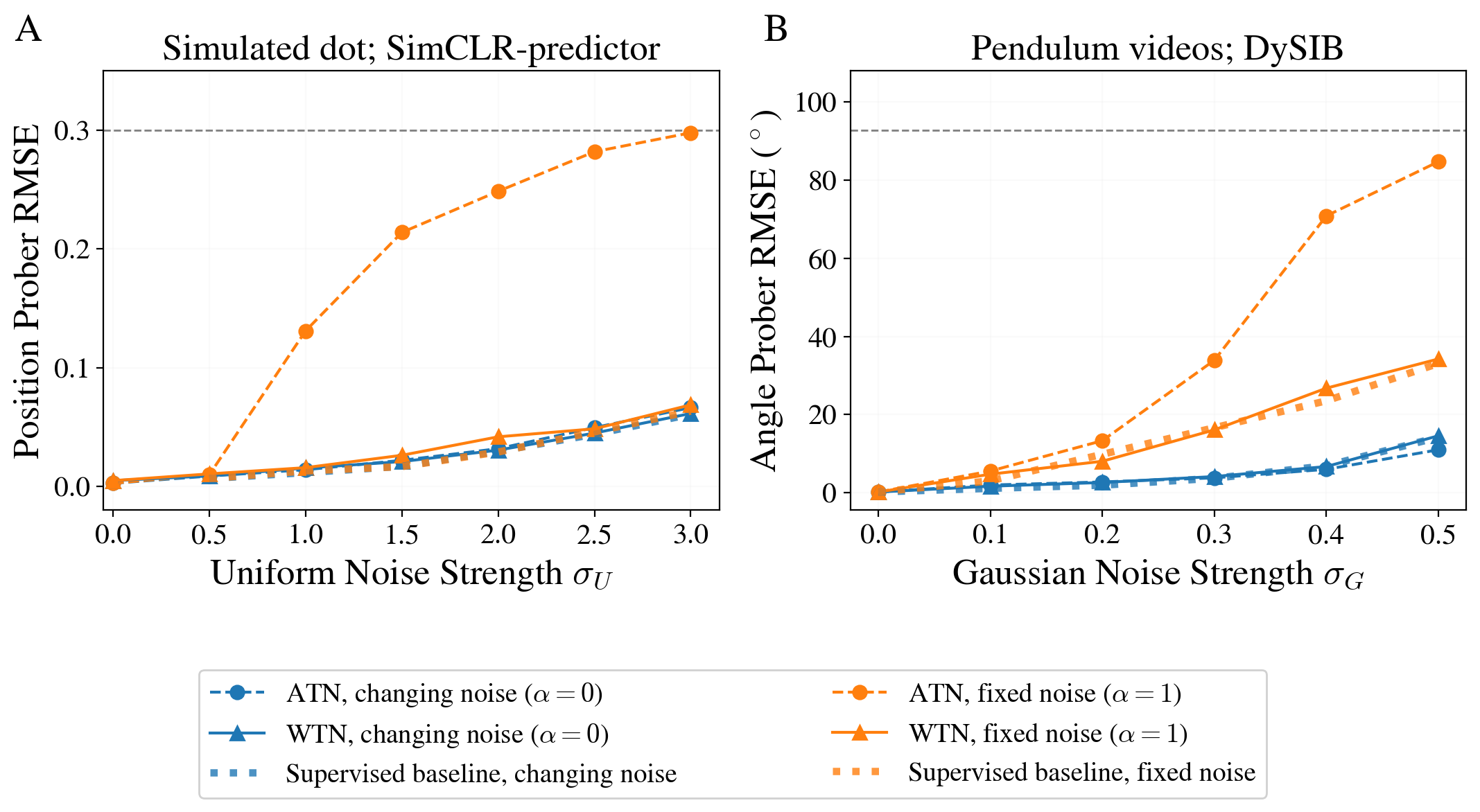}
    \caption{\textbf{ATN but not WTN predictive objectives fail under fixed noise.}
    (A) Stochastic moving-dot dataset with the SimCLR-predictor objective. (B) Experimental pendulum videos with the DySIB objective at $k_z=8$ (chosen following the original DySIB study to reduce sensitivity to local optima during training; Appendix~\ref{app:dataset_architectures}). We add temporally correlated noise to the observations and report the probe RMSE on the true dynamical variables as a function of noise strength $\sigma$. \emph{Changing noise} is sampled independently per frame. \emph{Fixed noise} is sampled once per trajectory and held constant within. Dotted lines show supervised baselines (one per noise regime, with the two coinciding on the moving dot). Thin dashed lines indicate performance under random guessing. Under ATN, the changing-noise probe error tracks its supervised baseline, while the fixed-noise probe error grows with $\sigma$ toward random guessing performance. Under WTN, both the changing- and the fixed-noise probe errors follow their respective supervised baseline curves. Training details are in Appendix~\ref{app:dataset_architectures}. See Appendix~\ref{sec:appendix-stability} for stability across hyperparameters and seeds.}
    \label{fig:RMSE_vs_sigma_changing_fixed_baseline}
\end{figure}

\section{Results}
\label{sec:results}
\subsection{ATN predictive objectives fail under fixed noise}\label{sec:atn-failure-rmse}

We first reproduce the known failure of ATN contrastive predictive objectives under fixed noise. The WTN variant is taken up in \S\ref{sec:wtn-results}. We begin with the supervised baselines. For the moving dot (Fig.~\ref{fig:RMSE_vs_sigma_changing_fixed_baseline}A), the encoder sees one frame at a time, so the temporal correlation of the noise is invisible and the changing- and fixed-noise baselines overlap. For the pendulum (Fig.~\ref{fig:RMSE_vs_sigma_changing_fixed_baseline}B), the supervised regressor takes two consecutive frames as input. With changing noise, the two frames carry independent noise realizations, so parts of the pendulum obscured in one frame are often visible in the other. With fixed noise, the same pattern is applied to both frames, providing no such redundancy. The fixed-noise baseline is therefore elevated even with full supervision.

On the moving dot, the standard ATN objective reproduces the failure reported by~\citet{sobal2022joint}. With changing noise, the SimCLR-predictor tracks the supervised baseline as $\sigma$ grows. With fixed noise, the probe error grows with $\sigma$ and saturates near random guessing, far above the (overlapping) baseline.
DySIB on the pendulum shows the same failure (Fig.~\ref{fig:RMSE_vs_sigma_changing_fixed_baseline}B). With changing noise, the predictor tracks the changing-noise baseline as $\sigma$ grows. With fixed noise, the probe error climbs with $\sigma$ toward random guessing, well above even the elevated fixed-noise baseline.
Despite very different compression schemes, both architectures fail similarly under fixed noise, showing that the failure is generic to contrastive predictive learning with across-trajectory negatives.

\begin{figure}[t]
    \centering
    \includegraphics[width=0.92\linewidth]{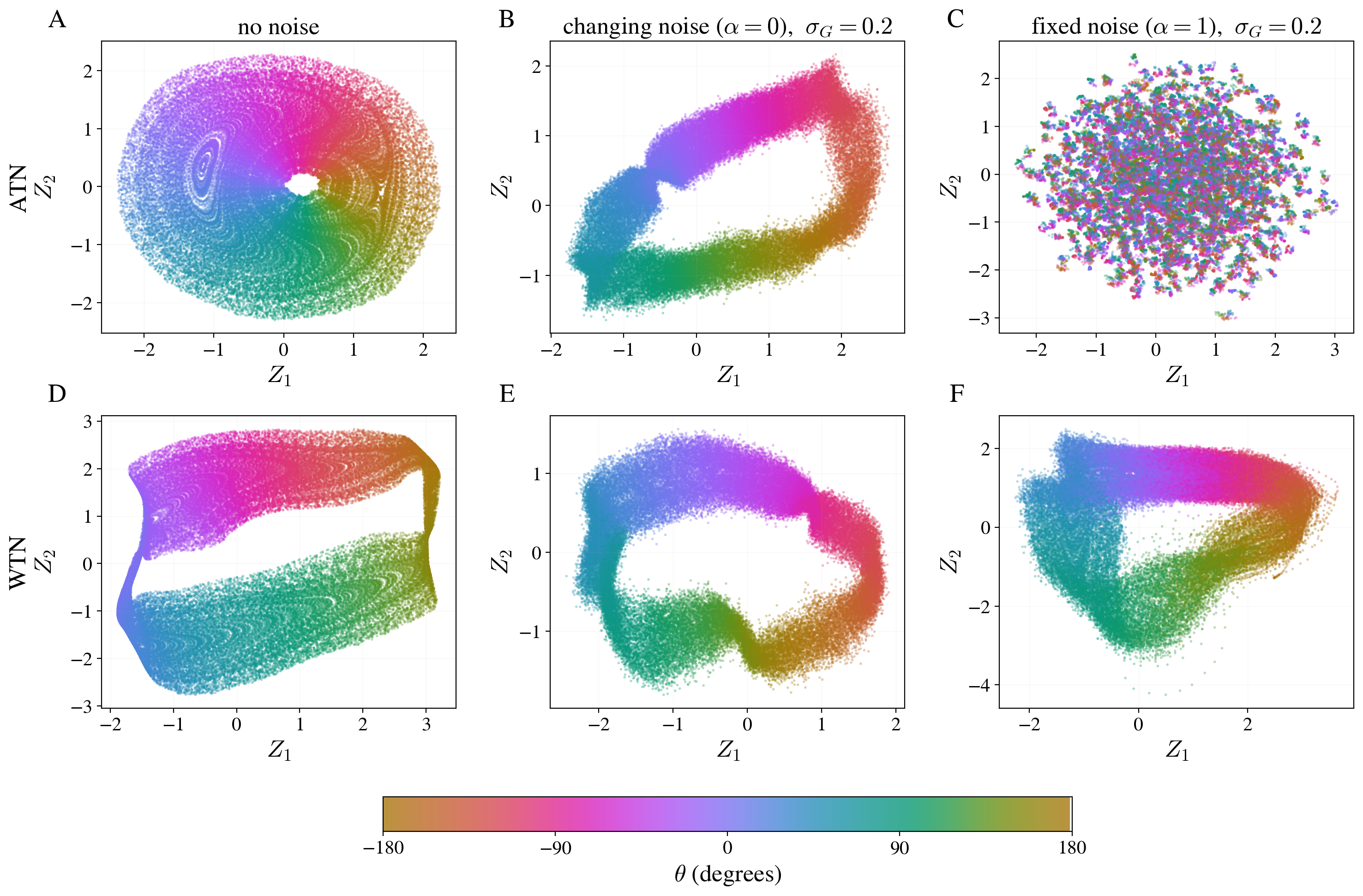}
     \caption{\textbf{Fixed noise breaks ATN but not WTN angular embeddings.}
    DySIB embeddings of the pendulum dataset with a two-dimensional bottleneck ($k_z=2$), colored by the true pendulum angle. Each point is a sample from the variational posterior $q(z_t \mid x_{t-n_F+1:t})$ for every $n_F$-frame block in the training data. Rows: ATN (top), WTN (bottom). Columns: no added noise, changing noise, fixed noise ($\sigma_G=0.2$ for both). Without added noise and with changing noise, both objectives produce latent spaces smoothly organized by angle. With fixed noise, the ATN representation organizes by trajectory-specific nuisances, breaking the global angular structure. WTN suppresses this predictive noise cue within the trajectory, and the fixed-noise embedding again organizes smoothly by angle.}
     \label{fig:embedding_baseline_dysib_kz_2}
\end{figure}

\subsection{Failure mode: ATN representations primarily encode trajectory-specific noise}\label{sec:atn-embedding-failure}

We would like to visualize the failure mode directly, which is feasible only with a low-dimensional latent. DySIB on the pendulum has $k_z=2$, matching the state dimensionality of angle and angular velocity, so its latent space can be inspected visually. The SimCLR-predictor on the moving dot uses 512-dimensional latents (Appendix~\ref{app:dataset_architectures}) and cannot be visualized the same way.

Figure~\ref{fig:embedding_baseline_dysib_kz_2} (top row) shows the latent embedding of the full dataset colored by the pendulum angle, not used for training. With no added noise (left column), the latent forms a ring whose angular coordinate tracks the physical angle. With changing noise (middle column), the ring is preserved, and a simple probe recovers the angle from the latent coordinates. With fixed noise (right column), the ring breaks into trajectory-specific clusters, and the dominant organization is now by trajectory-specific nuisance. Each trajectory still traces an arc whose direction reflects local angular motion (Fig.~\ref{fig:dysib_embedding_trajectory}), but the arcs are not aligned across trajectories, so the probe error is large (Fig.~\ref{fig:RMSE_vs_sigma_changing_fixed_baseline}).
This geometry is a consequence of the ATN loss: since different trajectories carry distinct fixed nuisances, nuisance encoding is a perfect way to separate negative pairs. 

\subsection{WTN objectives recover the underlying dynamical variables even with fixed noise}\label{sec:wtn-results}

Within-trajectory negatives (Eq.~\ref{eq:wtn}) fix the contrast structure by restricting the loss to compare each anchor only with other frames from the same trajectory. A nuisance that stays fixed within a trajectory takes the same value across all frames in the comparison, so it carries no information that distinguishes them and cannot reduce the loss. The encoder is therefore driven to encode features that vary along the trajectory, which are the true dynamical variables.

Figure~\ref{fig:RMSE_vs_sigma_changing_fixed_baseline} shows the WTN curves alongside the ATN curves analyzed earlier. With WTN, the fixed-noise probe error tracks the supervised baselines over the full range of noise strengths, on both the moving dot and the pendulum. The bottom row of Fig.~\ref{fig:embedding_baseline_dysib_kz_2} confirms this for the pendulum embedding: under fixed noise, the WTN representation again forms a ring-like structure organized by the true angle, with no trajectory-specific clustering.

WTN succeeds only when the training trajectories together cover a broad range of phase space, so that within-trajectory comparisons span the dynamics the encoder must represent (\S\ref{sec:setup}). To illustrate this, we now explore the dependence of the losses on $T$ and $\tau$.

\subsection{Dependence on noise strength and temporal correlation}\label{sec:tau-sweep}

\begin{figure}[t]
    \centering
    \includegraphics[width=0.85\linewidth]{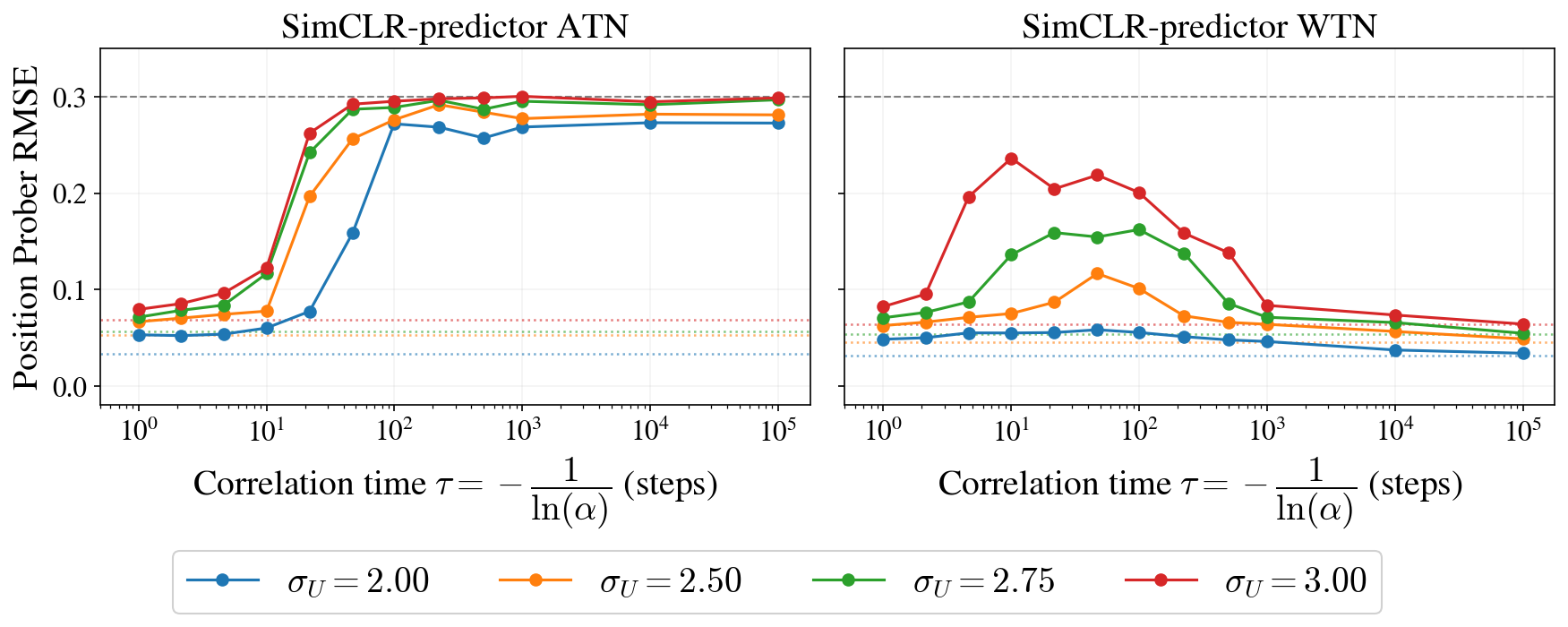}

    \caption{\textbf{After WTN removes the fixed-noise failure, the largest errors occur at intermediate noise correlation times.}
    Stochastic moving-dot dataset with temporally correlated uniform noise of strength $\sigma_U$ (colors) and correlation time $\tau$ (horizontal axis), at fixed trajectory duration $T=17$. Dotted colored horizontal lines mark the changing-noise ($\tau \to 0$) probe RMSE at the corresponding $\sigma_U$. Black dashed line is chance performance.
    (A) ATN. Error grows monotonically with $\tau$, saturating near random guessing in the fixed-noise limit. The crossover scales with $T$ (Fig.~\ref{fig:si_tau_transition_scaling}).
    (B) WTN. Error stays low at both ends and develops a peak at intermediate $\tau$. See Appendix~\ref{sec:appendix-stability} for stability across hyperparameters and seeds.}
    \label{fig:error_vs_tau_simclr_both}
\end{figure}

As we have seen, WTN removes the failure at fixed noise. We now explore what happens at intermediate correlation times. We use the OU noise model from the setup (\S\ref{sec:setup}), varying noise strength $\sigma_U$ and correlation time $\tau$ at fixed trajectory duration $T=17$ on the moving-dot dataset. Since in this and the next subsection vary $\tau$ and $T$, we omit the pendulum, whose maximum $T$ is fixed by the experimental videos, to conserve compute.

Figure~\ref{fig:error_vs_tau_simclr_both} shows that ATN probe error grows monotonically with $\tau$, saturating near random guessing in the fixed-noise limit. With WTN, error stays low at both ends and develops a finite peak at intermediate $\tau \approx T$. At those correlation times, the noise stays predictive within a trajectory yet varies enough across it so that WTN cannot treat it as a trajectory-level constant. This is expected since with matched correlation times, contrasts based on autocorrelation alone cannot separate noise from signal, connecting to Wiener's filtering~\citep{wiener1949matched} in classical signal processing. The peak height grows with $\sigma_U$, and is barely above the changing noise threshold ($\tau\rightarrow 0$) at small noise, $\sigma_U \lesssim 2$. Even at $\sigma_U = 1.5$ the noise is not negligible (Fig.~\ref{fig:datasets_noise}) and detecting the dot is hard. That WTN nevertheless filters out the noise across essentially all correlation times at this strength is surprising.

\subsection{Longer trajectories improve performance of WTN but not ATN objectives}
\label{sec:longer_trajectories}
We now explore the combined effect of trajectory duration $T$ and noise correlation time $\tau$, omitting the pendulum for the same reason as in \S\ref{sec:tau-sweep}. ATN is expected to remain near random guessing under fixed noise regardless of $T$, since longer videos reinforce the same trajectory-specific shortcut. For WTN, one expects that larger $T$ provides more within-trajectory training data, which the encoder can use to discriminate noise from signal even if $\tau \approx T$.

\begin{figure}
    \centering
    \includegraphics[width=0.9\linewidth]{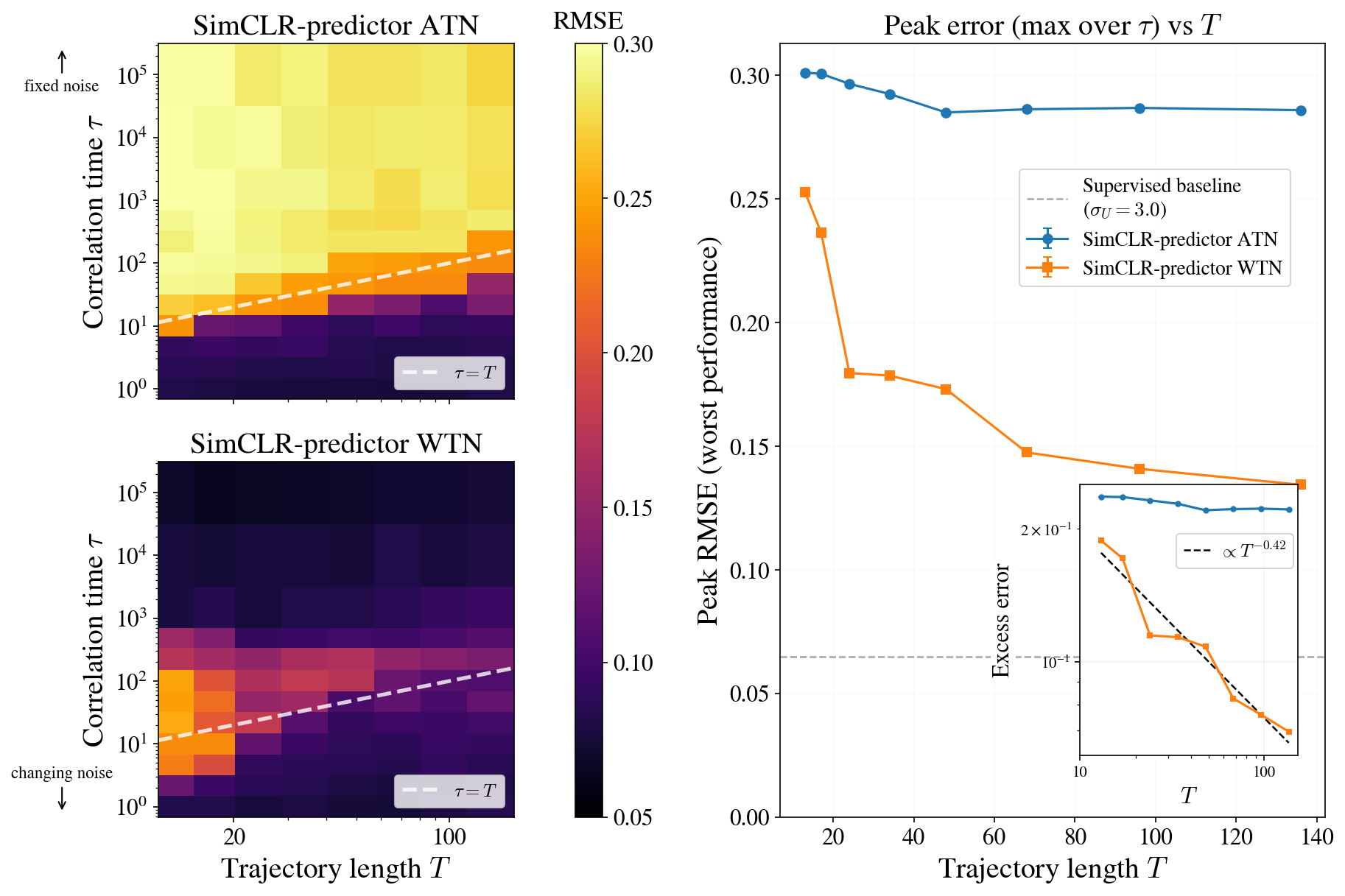}
    \caption{\textbf{Longer trajectories help WTN, not ATN.}
    Stochastic moving-dot dataset with temporally correlated uniform noise of strength $\sigma_U = 3.0$, varying trajectory duration $T$ and correlation time $\tau$.
    \textbf{(A)} Probe RMSE for ATN. The fixed-noise failure region at $\tau \gg T$ persists at every $T$.
    \textbf{(B)} Probe RMSE for WTN. The only remaining failure is the intermediate-correlation peak near $\tau \approx T$, which weakens as $T$ grows.
    \textbf{(C)} Maximum (worst) probe RMSE over $\tau$ as a function of $T$. ATN saturates near random guessing. WTN decreases toward the supervised baseline. Inset: excess of the WTN maximum over the supervised baseline, log--log axes. The decay is consistent with a power law in $T$. See Appendix~\ref{sec:appendix-stability} for stability across hyperparameters and seeds.}
        \label{fig:phase_T}
\end{figure}

Figure~\ref{fig:phase_T} demonstrates this for the stochastic moving-dot dataset with temporally correlated noise at fixed strength $\sigma_U = 3.0$ by sweeping $T$ and the correlation time $\tau$. For ATN, the failure region at $\tau \gg T$ persists at every $T$, and the maximum probe error over $\tau$ saturates near random guessing (Fig.~\ref{fig:phase_T}A,C). For WTN, the only remaining failure is the intermediate-$\tau$ peak around $\tau \approx T$, and it weakens monotonically as $T$ grows (Fig.~\ref{fig:phase_T}B). Expressed as excess over the supervised baseline, the WTN maximum decays as approximately a power law in $T$ (Fig.~\ref{fig:phase_T}C, inset), so WTN representations approach the supervised limit while ATN representations do not benefit from longer trajectories.

The error peak's location and shrinkage have different causes. As shown in \S\ref{sec:tau-sweep}, the peak is at $\tau \approx T$, where autocorrelation alone cannot separate signal from noise. It shrinks with $T$ because longer trajectories provide more data, letting the encoder utilize features beyond correlation time. Further, WTN also requires that the trajectory cover enough phase space for within-trajectory contrasts to be informative, and increasing $T$ helps. The moving dot covers phase space by construction; real-world trajectories confined to small regions may not, and would not gain from longer $T$ in the same way.

\section{Discussion}
\label{sec:discussion}
In two very different architectures (the deterministic SimCLR-JEPA and the variational information bottleneck of DySIB), we traced the common failure of encoders focusing on slow nuisances to the across-trajectory-negatives (ATN) structure of the contrastive objective. Under long-correlation-time noise, we showed that the encoder organizes its latent around the trajectory-specific nuisance, and probes cannot recover the dynamical variables. Replacing ATN with within-trajectory negatives (WTN) removes the predictive nuisance shortcut and recovers the latent variables relevant for dynamics. The residual probe error decreases with $T$ and peaks at $\tau \approx T$. Practitioners using contrastive predictive losses on real dynamical data should consider WTN as a default starting point when slow or intermediate-timescale nuisance is suspected.

\ourparagraph{Relation to prior work.}
The slow-noise failure is well documented across contrastive video SSL, where it has been called background bias, shortcut learning, or over-reliance on static features (see Introduction for citations). The standard remedies augment positive and negative pairs with image transformations that suppress the static content. We restructure the contrast itself, leaving the inputs unchanged. Conditioning pair selection on auxiliary structure has appeared in non-dynamic contexts, e.g., for fairness invariance~\citep{ma2021conditional}, but not as a remedy for slow-noise shortcuts in dynamics-paired predictive learning. 
The slow-nuisances failure traces back to objectives that take extracting slow features as their stated goal. For example, Slow Feature Analysis~\citep{wiskott2002slow} formalizes slowness as a learning principle, and Contrastive Predictive Coding (CPC)~\citep{oord2018representation} explicitly says ``these `slow features' that span many time steps are often more interesting.'' Thus ATN does what it was designed to do: under fixed noise, it encodes the slowest features available, which happen to be the trajectory-specific nuisance pattern. In other words, the premise that slowness is a useful proxy for predictive content must be re-evaluated.

\ourparagraph{Predictive sufficiency and nuisance invariance.}
\citet{achille2018emergence} characterize a good representation as sufficient for the task and minimal in everything else. With the future of a trajectory as the task, sufficiency means retaining the predictive information \citep{bialek2001predictability}, and minimality means discarding everything else. The contrast structure controls how the encoder approaches this. WTN holds the slow nuisance fixed across positives and negatives, removing it from the discriminative signal. The encoder must then encode within-trajectory variation, which, in physics language, is the dynamical state itself i.e. the same features across realizations that are predictive within each realization. ATN under fixed noise instead minimizes the loss via trajectory-specific nuisances, which distinguish negative pairs without predicting future dynamics.

\ourparagraph{Contrast as temporal filtering.}
WTN is a form of temporal filtering. By comparing frames within a trajectory, the loss is minimized by features that vary on the trajectory's timescale, neither too fast to average out nor too slow to be removed by WTN. This parallels the classical Wiener filtering~\citep{wiener1949matched}, where the filter that extracts a signal from noise depends on the autocorrelation structure of both. Thus when signal and noise share an autocorrelation time, contrasts have a hard time telling them apart, and the intermediate-$\tau$ peak in \S\ref{sec:tau-sweep} shows this. Resolving this problem requires using higher-order or spatial statistics in contrast design, such as contrasting delay-embedded latents at delays greater than one frame to expose higher-order temporal structures.

\ourparagraph{Objective design as conditional predictive information.}
The choice between WTN and ATN looks like a sampling technicality, but it can be interpreted as changing the quantity being optimized. As typically implemented, ATN draws negatives at the same time $t$ from other trajectories, so the InfoNCE bound is on $I(Z_t; \hat Z_t \mid t)$, the predictive mutual information averaged over realizations conditional on time. Any feature that varies between trajectories at fixed $t$ can tighten this bound, including a slow nuisance constant within a trajectory. WTN instead draws negatives from the same trajectory at other times, so the bound is on $I(Z_t; \hat Z_t \mid \text{trajectory})$, the predictive information that remains after conditioning on trajectory identity. Trajectory-level variables then cancel from the contrast, forcing the encoder to use only features that vary within a trajectory, the dynamical degrees of freedom, which are typically what matters physically. Neither choice gives the truly unconditional $I(Z_t; \hat Z_t)$. Estimating that would require sampling negatives jointly across time and trajectories, with care in re-weighting because trajectory lengths and counts are usually very uneven.

\ourparagraph{Computational efficiency.}
For batch size $N$ and trajectory length $T$, the number of negative comparisons scales as $\mathcal{O}(T N^2)$ for ATN and $\mathcal{O}(T^2 N)$ for WTN. When $N\gg T$, WTN is cheaper, though in practice $N$ is a tunable batch-size parameter. Thus WTN opens a useful tradeoff: fewer longer trajectories can become statistically and computationally competitive with many short ones.

\ourparagraph{Limitations.}
Our results come from two systems with known and low-dimensional dynamical states (synthetic moving-dot and experimental pendulum), with simple probe targets (dot position and pendulum angle). For systems with high-dimensional states or strongly chaotic dynamics, both the probe choice and the within-trajectory contrast may need more care. WTN also depends on each trajectory covering enough of the system's phase space; a stationary trajectory cannot resolve the dynamics regardless of $T$. With periodic orbits, when a trajectory spans more than one period, within-trajectory contrasts may push apart physically identical configurations, requiring $T$ restricted to one period. Even where coverage holds, the intermediate-$\tau$ peak persists at all $T$ we test, weakening at large $T$ but never disappearing, and we lack a theoretical account of its power-law decay. Finally, our analysis is restricted to InfoNCE-style contrastive losses; whether the same restructuring helps non-contrastive predictive losses is an open question.

\ourparagraph{Future work.}
The most direct next step is incorporating WTN into JEPA-style frameworks~\citep{balestriero2025lejepa,maes2026leworldmodel} and into predictive SSL on physical dynamics~\citep{martini2026dysib, chen2022automated}, where slow noise is the norm. More ambitiously, one may abandon explicit contrasts altogether, or learn the contrast itself, for example by attention mechanisms that select which past-future pairs to compare. A different approach could mix contrast designs, drawing negatives within and across trajectories on a schedule that encodes domain knowledge about which nuisances to suppress, in the spirit of motion-aware video contrasts~\citep{thoker2023tubelet}. Finally, in the spectral language of~\citet{ruizmorales2025koopman}, bare JEPA collapses onto eigenvalue-1 Koopman eigenfunctions, and whether different contrast designs move the encoder onto modes with $|\lambda| < 1$, recovering finite-timescale predictive content, is an open question.

\ourparagraph{Main message: Contrast design as inductive bias.}
The structure of the contrast determines what the encoder discards and what it preserves, selecting, in the language of \citet{achille2018emergence}, which point of the predictive-sufficient set the encoder converges to. Thus contrast design deserves attention on par with the architecture, the augmentations, or the form of the loss. For dynamical systems, the noise and the signal can have complex temporal structure, unknown a priori. Consequently, whether SSL on dynamics is feasible without substantial domain knowledge or expensive exploration of contrast designs is, in our view, the open question this work highlights.

\section*{Acknowledgments}
We would like to thank Randall Balestriero, Ravid Shwartz-Ziv, K.\ Michael Martini, and Eslam Abdelaleem for useful discussions. This work was funded, in part, by the Simons Foundation Investigator grant to IN and by the National Science Foundation Grant No.~2409416. PG was additionally funded by the Tarbutton Interdisciplinary Postdoctoral Fellowship at the Emory College of Arts and Sciences.  We acknowledge support of our work through the use of the HyPER C3 cluster of Emory University’s AI.Humanity Initiative.


\bibliographystyle{plainnat}   
\bibliography{ref}

\newpage
\appendix

\setcounter{figure}{0}
\renewcommand{\thefigure}{S\arabic{figure}}

\setcounter{table}{0}
\renewcommand{\thetable}{S\arabic{table}}

\etocdepthtag.toc{atoc}

\etocsettagdepth{mtoc}{none}
\etocsettagdepth{atoc}{subsection}

\etocsettocstyle{\section*{Appendix: Table of Contents}}{}
\tableofcontents

\section{Supplementary results}
\subsection{Learned pendulum representations in two dimensions}

\begin{figure}[h]
    \centering
    \includegraphics[width=0.8\linewidth]{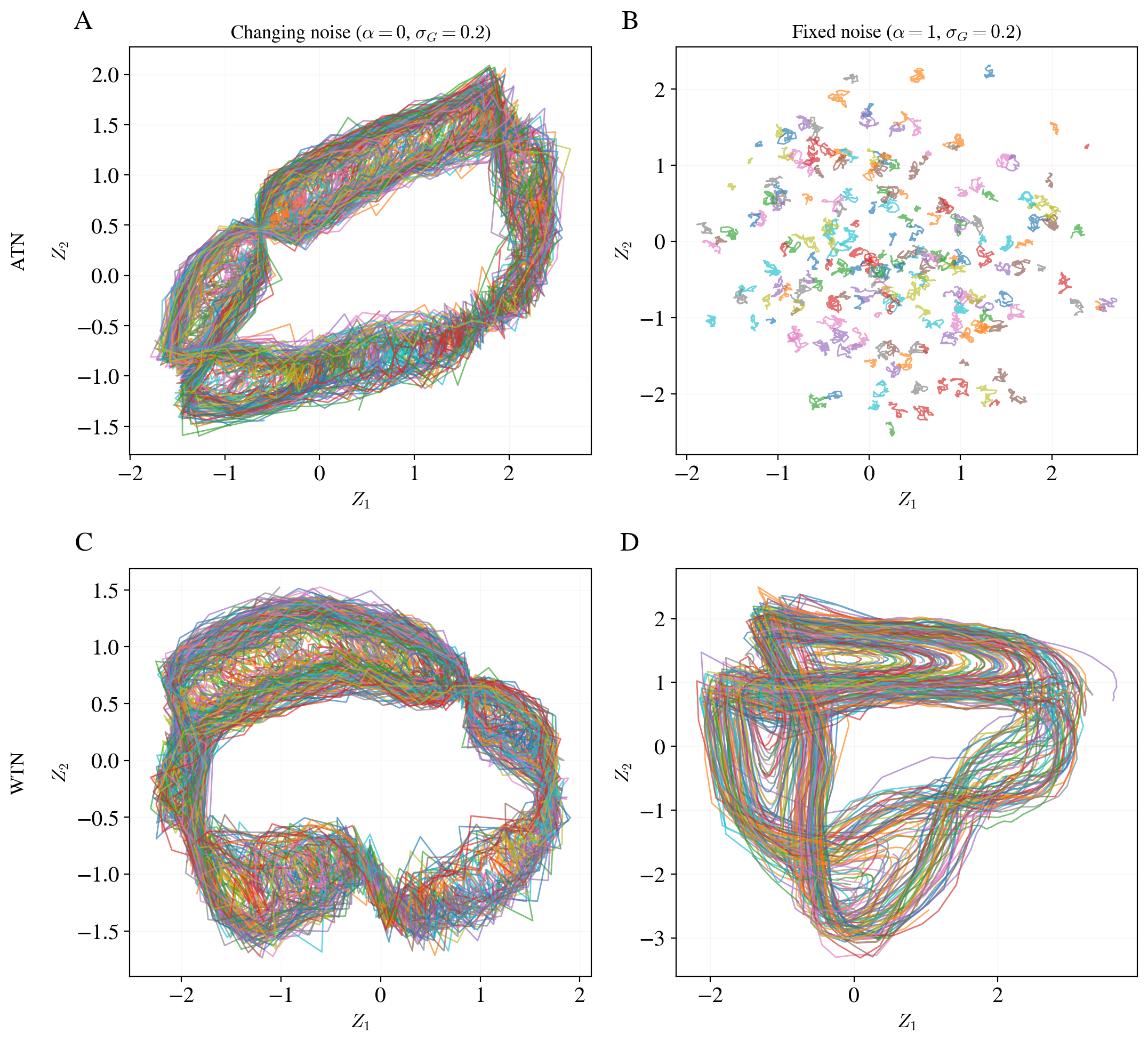}
    \caption{\textbf{Trajectory-connected latent paths reveal ATN's clustering by trajectory identity under fixed noise.} Latent-space trajectories of individual test-set pendulum sequences in the 2-D learned representation ($k_z=2$). Rows: ATN (top), WTN (bottom). Columns: changing noise ($\alpha=0$, $\sigma_G=0.2$; panels A, C) and fixed noise ($\alpha=1$, $\sigma_G=0.2$; panels B, D). Same trained representations as in Fig.~\ref{fig:embedding_baseline_dysib_kz_2}, now with points connected within each trajectory and colored by trajectory index.}
    \label{fig:dysib_embedding_trajectory}
\end{figure}
\FloatBarrier

DySIB with across-trajectory negatives (ATN) on pendulum data does not smoothly encode the relevant dynamical variable (angle) under fixed noise (Fig.~\ref{fig:embedding_baseline_dysib_kz_2}). Figure~\ref{fig:dysib_embedding_trajectory} shows the held-out test-set embeddings, with points connected within each trajectory and colored by trajectory index (colors repeat across the test set). As discussed in the main text, both ATN and WTN disregard trajectory identity under changing noise. Under fixed noise, ATN instead organizes the learned embeddings around trajectory identity, i.e., the fixed noise patterns; WTN removes this predictive shortcut.

\subsection{Stability to hyperparameter choices and random seeds}\label{sec:appendix-stability}

\begin{figure}[h]
    \centering
    \includegraphics[width=\linewidth]{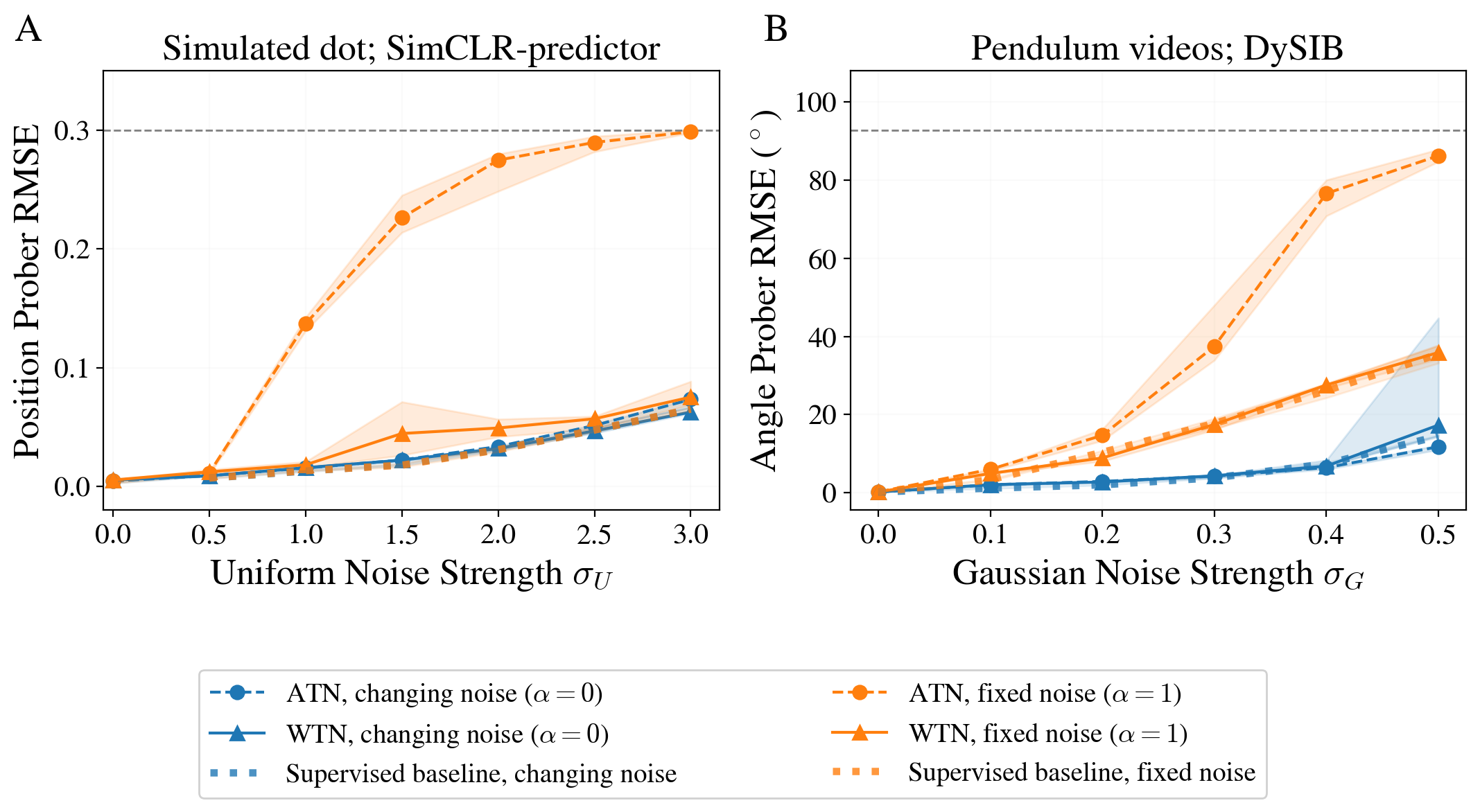}
    \caption{\textbf{Stability of the noise-strength sweep to hyperparameter choices and random seeds.} Prober RMSE for fixed and changing noise in both datasets as a function of noise strength; parallels Fig.~\ref{fig:RMSE_vs_sigma_changing_fixed_baseline}. (A) Moving dot: 50 pseudorandom hyperparameter combinations per point. Light fill spans min--max probe RMSE over the 5 trials with the smallest error; lines and points mark the median. (B) Pendulum: 15 replicate seeds per point with fixed hyperparameters; same fill/median convention applied to the 5 trials with the lowest error.}
    \label{fig:si_noise_strength_hp}
\end{figure}

\begin{figure}[h]
    \centering
    \includegraphics[width=\linewidth]{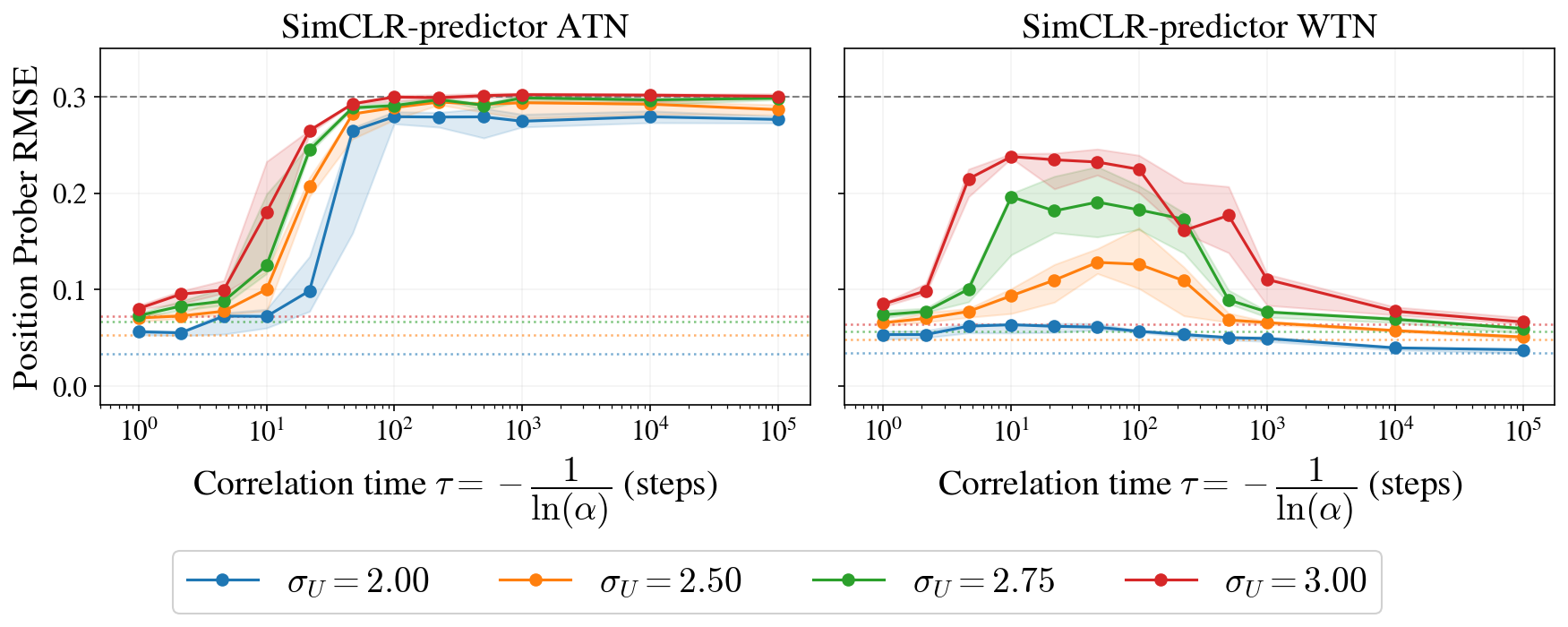}
    \caption{\textbf{Hyperparameter variability of the $\tau$-sweep.} Prober RMSE for the moving dot using the SimCLR-predictor as a function of noise correlation time $\tau$, for both ATN and WTN; parallels Fig.~\ref{fig:error_vs_tau_simclr_both}. In both panels, 50 pseudorandom hyperparameter combinations per point; light fill spans min--max probe RMSE over the 5 trials with the smallest error; lines and points mark the median.}
    \label{fig:si_tau_hp}
\end{figure}

For the SimCLR-predictor on the moving-dot dataset, we train the model with 50 pseudorandom hyperparameter combinations spanning orders of magnitude for each noise and dataset configuration $(\sigma, \tau, T)$ (see Appendix~\ref{app:dataset_architectures} for training details). Throughout the main text, we report the minimum probe RMSE over all 50 trials. To show that this is not an artifact of a single lucky hyperparameter setting, Figs.~\ref{fig:si_noise_strength_hp} and~\ref{fig:si_tau_hp} reproduce the main-text plots using the 5 best hyperparameter combinations: the light fill spans the range of probe RMSEs over those 5 trials, and lines and scatter points mark the median. We do not show mean and standard deviation, since averaging over different hyperparameter values is not meaningful. For the pendulum panel (Fig.~\ref{fig:si_noise_strength_hp}B), we instead vary the random seed at fixed hyperparameters: 15 replicate seeds per point, with the same fill/median convention applied to the 5 trials with the smallest error. The two protocols probe variability under two distinct sources of stochasticity, hyperparameter choice for the synthetic moving-dot setting and seed-level randomness for the experimental pendulum, each natural for its setting.

\FloatBarrier
\subsection{Scaling with trajectory duration $T$}
This section supplements \S\ref{sec:longer_trajectories}.

\begin{figure}[h]
    \centering
    \includegraphics[width=0.9\linewidth]{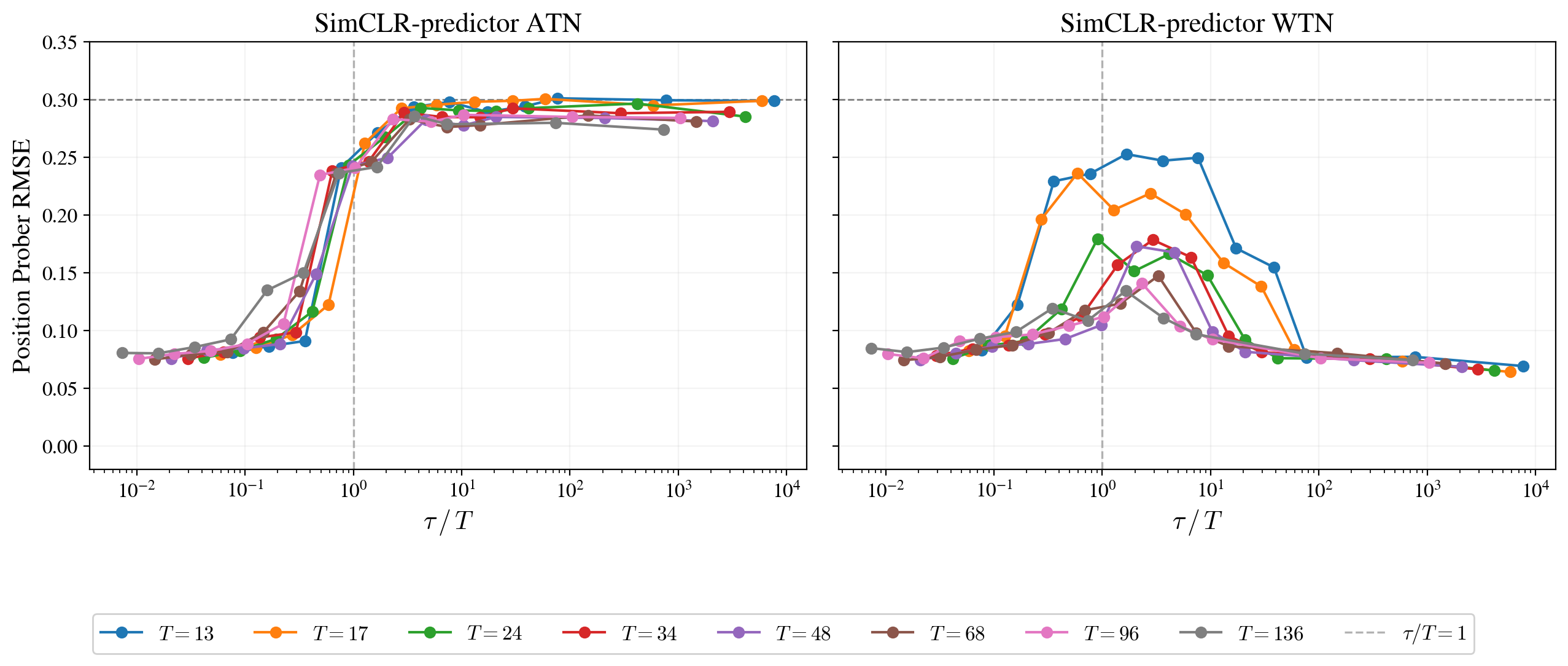}
    \caption{\textbf{Probe error collapses under $\tau/T$ rescaling for ATN, while WTN's peak shifts with $T$.} Probe RMSE for the SimCLR-predictor with (A) ATN and (B) WTN, plotted against the rescaled correlation time $\tau/T$ at fixed noise amplitude $\sigma_U=3.0$. Each colored curve is a different trajectory duration $T$.}
    \label{fig:si_tau_transition_scaling}
\end{figure}

Figure~\ref{fig:si_tau_transition_scaling} plots the probe error from the phase diagrams of Fig.~\ref{fig:phase_T}A,B against the rescaled correlation time $\tau/T$. Under ATN, the rescaled data collapse with a transition at $\tau/T \approx 1$, consistent with the phase diagram in the main text. Longer trajectories do not remove the fixed-noise shortcut: any finite-length trajectory retains this structure in the representation. Under WTN, the peak error decays as $T$ grows, with the peak position remaining at $\tau \approx T$.

\begin{figure}[h]
    \centering
    \includegraphics[width=0.9\linewidth]{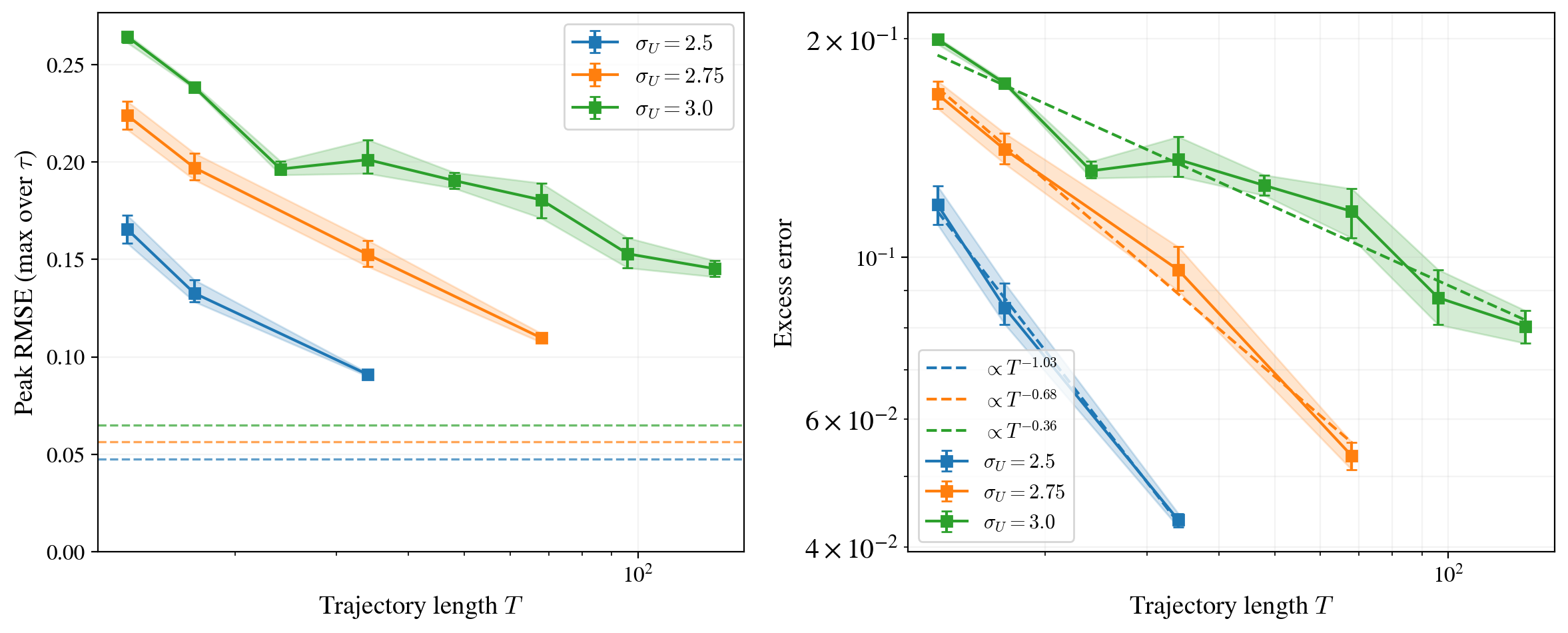}
    \caption{\textbf{WTN's peak error decays as a power law in $T$ across noise strengths.} Peak probe RMSE as a function of trajectory length $T$ for the SimCLR-predictor with WTN at different noise magnitudes $\sigma_U$. (A) Absolute peak RMSE (maximum over $\tau$). Dashed horizontal lines mark the supervised baseline floor at each $\sigma_U$. (B) Excess peak RMSE above the supervised floor, log--log axes. Dashed lines show power-law fits $\sim T^{-\gamma}$. Shaded regions and error bars are $68\%$ bootstrap confidence intervals ($N=500$ resamples). The excess error follows a power law in $T$ at all three noise strengths. The exponent's magnitude is larger at smaller noise, so the problem is easier and scales more favorably with $T$ as the noise weakens.}
    \label{fig:si_power_law_scaling_sigmas}
\end{figure}

Figure~\ref{fig:si_power_law_scaling_sigmas} extends Fig.~\ref{fig:phase_T}C, which shows the decay of WTN's peak error at $\sigma_U=3.0$, to smaller $\sigma_U$, where the overall error at $T=17$ is already smaller (cf.\ Fig.~\ref{fig:error_vs_tau_simclr_both}). The excess error (peak error minus supervised-baseline floor) decays approximately as a power law in $T$ for each $\sigma_U$, with faster decay at smaller noise strengths. We do not claim a universal exponent and leave this for future investigation.

\FloatBarrier

\section{Datasets, architectures and training}\label{app:dataset_architectures}

\subsection{Dataset details}
\begin{figure}[h]
    \centering
    \includegraphics[width=\linewidth]{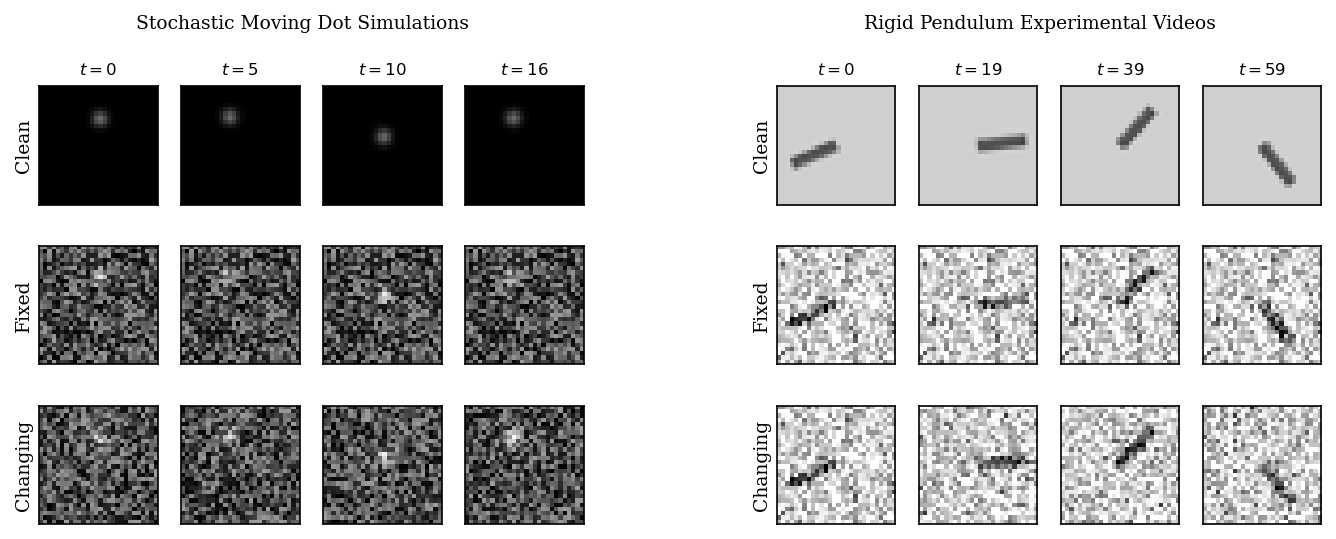}
    \caption{\textbf{Example frames from the two datasets.} \textbf{Left:} moving-dot dataset~\citep{sobal2022joint}, noise amplitude $\sigma_U = 1.5$. \textbf{Right:} pendulum dataset~\citep{chen2022automated}, noise amplitude $\sigma_G = 0.2$. \textbf{Top row:} clean noiseless frames. \textbf{Middle row:} fixed noise. \textbf{Bottom row:} changing noise.}
    \label{fig:datasets_noise}
\end{figure}

We use two datasets, both illustrated in Fig.~\ref{fig:datasets_noise}. The moving-dot dataset~\citep{sobal2022joint} is synthetic: each trajectory follows a single point moving in a unit square under random actions $a_t \in \mathbb{R}^2$ of bounded norm $\|a_t\|_2 \leq 0.14$, with the dot's position clipped to remain inside the square. Action directions are sampled from a von Mises distribution centered on a per-trajectory global angle (concentration $1.3$), and step sizes are sampled uniformly on $[0, 0.14]$, which gives each trajectory a randomly chosen drift direction and ensures non-zero expected displacement. The rendered frames $x_t$ are $28 \times 28$ grayscale images ($D = 784$) showing the dot blurred by a Gaussian of width $0.05$, and trajectories have length $T = 17$ frames (the initial frame plus $16$ actions), matching the default in~\citet{sobal2022joint}; we additionally vary $T$ in some experiments to study the dependence of the learned representation on trajectory duration. Since this dataset is synthetic, fresh trajectories are generated for both training and evaluation.

The pendulum dataset~\citep{chen2022automated} consists of real experimental video recordings of a rigid pendulum swinging under gravity. Each video is downsampled to $28 \times 28$ grayscale ($D = 784$) frames, with pixel values normalized to $[0, 1]$, and contains $T = 60$ frames; we use $1000$ videos for training and $200$ held out for testing, following~\citet{martini2026dysib}. 

In both cases, noise with different strengths and temporal correlations is added to the frames as described in the main text.

\subsection{SimCLR JEPA architecture and training}
\label{sec:simclr_arch_training}
We use the SimCLR-predictor of~\citet{sobal2022joint} unchanged, available under the MIT license. The encoder $g$ has three 2D convolutional blocks with channel counts $(32, 64, 64)$, kernel sizes $(5, 5, 3)$, strides $(2, 2, 1)$, and padding $(2, 2, 1)$; each block is followed by a ReLU nonlinearity and BatchNorm. A $2 \times 2$ average pool and a single fully-connected layer then produce the 512-dimensional latent $z_t = g(x_t) \in \mathbb{R}^{512}$. The predictor $f$ is a recurrent network with one GRU layer of width 512 driven by the action input $a_t \in \mathbb{R}^2$, so that $\hat z_1 = z_1 = g(x_1)$ initializes the GRU hidden state, and subsequent applications produce $\hat z_t = f(\hat z_{t-1}, a_{t-1})$ for $t = 2, \ldots, T$, with the GRU hidden state at each step taken as the predicted latent.

The training objective is the symmetrized InfoNCE used for JEPA in~\citet{sobal2022joint},
\begin{equation}
    \mathcal{L}_{\rm SimCLR-JEPA} = \frac{1}{2(T-1)} \sum_{t=2}^{T} \left[ \mathrm{InfoNCE}(\hat z_t, z_t) + \mathrm{InfoNCE}(z_t, \hat z_t) \right],
    \label{eq:simclr_jepa_objective}
\end{equation}
where each $\mathrm{InfoNCE}(\cdot, \cdot)$ takes the form of Eq.~\eqref{eq:infonce} averaged over the batch, with similarity score $s(u, v) = u^\top v / \tau_{\rm NCE}$ (applied to L2-normalized or unnormalized vectors depending on the condition; see below) at temperature $\tau_{\rm NCE}$. For the objective with across-trajectory negatives (ATN), negatives are drawn from all other trajectories in the batch at each time step; for the objective with within-trajectory negatives (WTN), negatives are all other time steps within the same trajectory. The loss is computed directly on the 512-dimensional encoder outputs with an identity projection head. Frames are not augmented, and the encoder and predictor are trained jointly for 100 epochs using the LARS optimizer with a cosine learning-rate schedule and a 10-epoch linear warm-up.

Four hyperparameters are selected per noise condition $(\alpha, \sigma, T)$ by random search: LARS base learning rate $\eta_{\rm base}$, log-spaced over 16 values from $2.5\times10^{-4}$ to $3.9$ and scaled with batch size as $\eta = \eta_{\rm base}\times B/256$; InfoNCE temperature $\tau_{\rm NCE}$, log-spaced over 9 values from $0.1$ to $128$; batch size $B \in \{64, 128, 256, 512\}$; and whether encoder outputs are L2-normalized before computing the loss. For each noise condition we draw 50 random combinations from the Cartesian product of these grids and report the lowest probe RMSE in the main text. Prober architecture and training are described below.

\subsection{DySIB architecture and training}
We use the DySIB architecture and training setup of~\citet{martini2026dysib} unchanged, available under the MIT license. The window encoder $\phi$ is built from a single per-frame three-layer fully connected MLP with hidden width 256 and ReLU activations, applied to each of the $n_F = 2$ frames. The per-frame embeddings are concatenated and fed to two parallel linear heads producing the mean $\mu \in \mathbb{R}^{k_z}$ and log-variance $\ell \in \mathbb{R}^{k_z}$ of the Gaussian variational posterior $q(z_t \mid x_{t-n_F+1}, \ldots, x_t) = \mathcal{N}(\mu, \mathrm{diag}(\exp \ell))$. The same backbone and heads are shared between the past and future windows. The $\delta$-predictor is a three-layer fully connected MLP with hidden width 64 and ReLU activations, producing both the shift $\delta(z_t)$ that defines $\hat z_{t+n_F} = z_t + \delta(z_t)$ and a per-dimension log-variance for the predictive distribution.

The DySIB training objective~\citep{martini2026dysib} is the symmetric variational information bottleneck
\begin{equation}
\mathcal{L}_{\rm DySIB} = \widetilde{I}^E(X; Z_t) + \widetilde{I}^E(Y; Z_{t+n_F}) - \beta\,\widetilde{I}^D(Z_t; Z_{t+n_F}),
\label{eq:dysib_objective}
\end{equation}
where $X$ and $Y$ denote the past and future frame windows, and the superscripts $E$ and $D$ stand for encoder-side (compression) and decoder-side (prediction) terms. The encoder-side $\widetilde{I}^E$ terms are upper bounds on $I(X; Z_t)$ and $I(Y; Z_{t+n_F})$, implemented as KL divergences~\citep{alemi2017deep} between each posterior and a standard normal prior. The decoder-side $\widetilde{I}^D$ term is an InfoNCE lower bound on $I(Z_t; Z_{t+n_F})$, derived as in~\citet{abdelaleem2025deep}, with critic given by the log-density $s(z_t, z_{t+n_F}) = \log r(z_{t+n_F} \mid z_t)$ of the Gaussian predictive distribution $r(\cdot \mid z_t) = \mathcal{N}(z_t + \delta(z_t),\, \mathrm{diag}(\exp \ell_\delta(z_t)))$, where $\ell_\delta(z_t)$ denotes the per-dimension log-variance head of the $\delta$-predictor. As with the SimCLR-predictor (\S\ref{sec:simclr_arch_training}), the InfoNCE term is typically evaluated with across-trajectory negatives (ATN); here we also explore within-trajectory negatives (WTN). We set $\beta = 100$, large enough that the compression terms contribute negligibly relative to the predictive one. The latent dimensionality $k_z$ then controls the effective bottleneck, not the compression weight. All weight matrices use Xavier-uniform initialization, and we train with Adam at learning rate $10^{-4}$, batch size $1024$, for $100$ epochs.

The latent dimension $k_z$ used in this work varies across analyses. We use $k_z = 2$ for embedding visualizations (Fig.~\ref{fig:embedding_baseline_dysib_kz_2}), where the two-dimensional latent matches the two phase-space degrees of freedom of the pendulum, the angle and the angular velocity. For prober experiments we use $k_z = 8$. As shown in~\citet{martini2026dysib}, the predictive mutual information $I(Z_t; Z_{t+n_F})$ saturates at $k_z = 2$, so latents at any $k_z \geq 2$ retain equivalent predictive information. Slightly enlarging the latent space reduces the fraction of optimization runs that fail to find a good local optimum, while the effective intrinsic dimension of the learned manifold remains close to two. For details about prober architecture and training, see below.

\subsection{Prober architectures and training}
Probers are minimal models that map the frozen learned representation to known dynamical targets, kept low-capacity so that probe accuracy reflects the representation rather than the probe's own learning.

For the moving-dot dataset, the SimCLR-predictor latent is 512-dimensional and, matching~\citet{sobal2022joint}, the prober is a linear map (no bias) from this latent to the 2-dimensional dot position, trained with Adam (learning rate $10^{-3}$) for 30 epochs using an MSE loss, with the SimCLR weights frozen. The latent is high-dimensional enough that a linear probe suffices.

For the pendulum, the targets are nonlinear functions of the DySIB latent (Fig.~\ref{fig:embedding_baseline_dysib_kz_2} illustrates the case $k_z = 2$), so a prober that can capture nonlinear structure in a low-dimensional setting is needed. We use a random-feature linear prober~\citep{rahimi2007random}: a single fixed (randomly initialized, never updated) linear projection from $k_z$ to 512 dimensions (matching the dimensionality of the embedding for the moving dot) followed by a Softplus nonlinearity, with a \emph{trainable linear readout}. The expansion weights are drawn from $\mathcal{N}(0,2/\sqrt{k_z})$, the variance-preserving initialization for Softplus. The angle $\theta$ is regressed in $(\cos\theta, \sin\theta)$ form, with predictions converted back via $\mathrm{atan2}$ and angular error wrapped to handle the $\pm\pi$ boundary. The readout is fitted, with DySIB weights frozen, by closed-form ridge regression (regularization $\lambda = 10^{-3}$). During DySIB training, a fresh prober is fitted from scratch every epoch on the current model's frozen representations (a single closed-form solve), yielding a trajectory of test $\theta$-RMSE values. The reported test RMSE is the minimum over this trajectory, i.e., an oracle estimate of the best representation the model achieves at any point during training, which guards against the representation degrading in the final epochs. We verified that the qualitative conclusions are unchanged when using the final-epoch prober instead.

\subsection{Computational resources}
\label{sec:compute_resources}
SimCLR experiments were conducted on AWS instances of NVIDIA B200 GPUs. As a representative single run at the canonical $T = 17$ and batch size 256, with a linear position prober (30 training epochs) evaluated every 10 epochs of encoder training, runtime was approximately 120\,s for the within-trajectory-negative variant (WTN) and 200\,s for the across-trajectory-negative variant (ATN). The bulk of compute arose from the hyperparameter sweeps, which spanned $13$ OU memory values $\alpha$, $4$ noise amplitudes $\sigma$ at $T = 17$ for both variants, $7$ additional trajectory lengths $T$ for $\sigma = 3.0$ for both variants, and (see Fig.~\ref{fig:si_power_law_scaling_sigmas}) $2$ and $3$ further trajectory lengths for $\sigma = 2.5$ and $\sigma = 2.75$ respectively for the WTN variant, amounting to 351 distinct $(T,\sigma,\alpha,\text{variant})$ conditions, each evaluated with 50 random hyperparameter combinations (as described in Sec.~\ref{sec:simclr_arch_training}), for approximately $17,500$ training runs in total. At an average of ${\sim}3$ min per run (the compute time per run depends on the trajectory length, batch size and negative variant), the total compute required approximately 900 GPU-hours.

DySIB experiments were conducted on AWS instances of NVIDIA H100 GPUs. Each run trained the model for 100 epochs with a fresh prober (50 training epochs) evaluated every two epochs of DySIB training; including this probing overhead, each run took approximately 10 minutes. The full paper sweep comprised $2\;(k_z) \times 6\;(\sigma) \times 2\;(\alpha) \times 2\;(\text{WTN/ATN variant}) \times 15\;(\text{replicates}) = 720$ independent runs, totalling approximately 120 GPU-hours.

\newpage




\end{document}